%% file: main.tex
\documentclass[10pt,twocolumn,letterpaper]{article}

\usepackage{cvpr}
\usepackage{times,fullpage,times}
\usepackage{url}
\usepackage{epsfig}
\usepackage{graphicx}
\usepackage{amsmath}
\usepackage{amssymb}
\usepackage{caption}
\usepackage{subcaption}
\usepackage{algorithm}
\usepackage{algpseudocode}
\usepackage{pifont}
\usepackage{xcolor}
\usepackage{rotating}
\usepackage{epstopdf}
\usepackage{caption}
\usepackage{subcaption}\usepackage{multirow}

\input{math_definition.tex}
\usepackage[pagebackref=true,breaklinks=true,letterpaper=true,colorlinks=true,bookmarks=false]{hyperref}
\usepackage{bibspacing}

\cvprfinalcopy 

\def\cvprPaperID{1957} 

\ifcvprfinal\fi
\begin{document}

\title{Summary Transfer: Exemplar-based Subset Selection for Video Summarization}

\author{Ke Zhang$^*$, Wei-Lun Chao\thanks{\hspace{4pt}Equal contributions}\\
University of Southern California\\
Los Angeles, CA 90089\\
{\tt\small \{zhang.ke,weilunc\}@usc.edu}
\and
Fei Sha\\
University of California\\
Los Angeles, CA 90095\\
{\tt\small feisha@cs.ucla.edu}
\and
Kristen Grauman\\
University of Texas at Austin\\
Austin, TX 78701\\
{\tt\small grauman@cs.utexas.edu}
}

\maketitle

\begin{abstract}
\input{abstract}
\end{abstract}

\input{intro}

\input{related}
\input{approach}

\input{experiment}

\input{conclusion}
\section*{Acknowledgements}
KG is partially supported by NSF IIS - 1514118. Others are partially supported by USC Annenberg and Viterbi Graduate Fellowships, NSF IIS - 1451412, 1513966, and CCF-1139148.
\setlength{\bibitemsep}{.01\baselineskip}
{\footnotesize
\bibliographystyle{ieee}
\bibliography{main}
}
\clearpage
\section*{Supplementary Material}
\appendix
In this Supplementary Material, we give out details omitted in the main text: learning parameters in section~\ref{sMLE} (section~3.3 in the main text), datasets, features and evaluation metrics for our experimental studies in section~\ref{sData}, and \ref{sEval} (section~4.1 in the main text), analysis and results in section~\ref{sRes} (section~4.2 in the main text), and further discussions in section~\ref{sdDisc}.

\input{gradient}

\input{datasets}
\input{eval_metric}
\input{supp_exp}

\end{document}

%% file: math_definition.tex
\usepackage{amsmath,amsfonts}
\usepackage{amssymb,amsopn}
\usepackage{bm} 

\usepackage{amssymb}
\usepackage{amsmath,amsfonts}
\usepackage{amsthm,amsopn}

\usepackage{bm} 

\newcommand{\vct}[1]{\boldsymbol{#1}} 
\newcommand{\mat}[1]{\boldsymbol{#1}} 
\newcommand{\cst}[1]{\mathsf{#1}}  

\newcommand{\T}{^{\textrm T}} 

\newcommand{\twonorm}[1]{\left\|#1\right\|_2^2}

\newcommand{\ProbOpr}[1]{\mathbb{#1}}

\newcommand{\expect}[2]{%
\ifthenelse{\equal{#2}{}}{\ProbOpr{E}_{#1}}
{\ifthenelse{\equal{#1}{}}{\ProbOpr{E}\left[#2\right]}{\ProbOpr{E}_{#1}\left[#2\right]}}} 
\newcommand{\var}[2]{%
\ifthenelse{\equal{#2}{}}{\ProbOpr{VAR}_{#1}}
{\ifthenelse{\equal{#1}{}}{\ProbOpr{VAR}\left[#2\right]}{\ProbOpr{VAR}_{#1}\left[#2\right]}}} 

\DeclareMathOperator{\argmax}{arg\,max}

\newcommand{\ground}{{\mathcal{Y}}} 

\newcommand{\valpha}{\vct{\alpha}}

\newcommand{\vy}{\vct{y}}

\newcommand{\vv}{\vct{v}}

\newcommand{\mU}{\mat{U}}

\newcommand{\mS}{\mat{S}}

\newcommand{\mM}{\mat{M}}

\newcommand{\cN}{\cst{N}}

\newcommand{\cR}{\cst{R}}

\newcommand{\mC}{\mat{C}}

\newcommand{\mL}{\mat{L}}
\newcommand{\mI}{\mat{I}}

\newcommand{\vphi}{\vct{\phi}}
\newcommand{\mOmega}{\mat{\Omega}}

\newcommand{\eat}[1]{}

%% file: abstract.tex
Video summarization has unprecedented importance to help us digest, browse, and search today's ever-growing video collections. We propose a novel subset selection technique that leverages supervision in the form of human-created summaries to perform automatic keyframe-based video summarization. The main idea is to nonparametrically transfer summary structures from annotated videos to unseen test videos.   We show how to extend our method to exploit semantic side information about the video's category/genre to guide the transfer process by those training videos semantically consistent with the test input.  We also show how to generalize our method to subshot-based summarization, which not only reduces computational costs but also provides more flexible ways of defining visual similarity across subshots spanning several frames.  We conduct extensive evaluation on several benchmarks and demonstrate promising results, outperforming existing methods in several settings.

%% file: intro.tex
\section{Introduction}
\label{sIntro}

The amount of video data has been explosively increasing due to the proliferation of video recording devices such as mobile phones, wearable and ego-centric cameras, surveillance equipment, and others. According to YouTube statistics, about 300 hours of video are uploaded every minute~\cite{Youtube_url}. To cope with this video data deluge, \emph{automatic video summarization} has emerged as a promising tool to assist in curating video contents for fast browsing, retrieval, and understanding \cite{caba2015activitynet, lu2016coherent, wu2016Harnessing, Yang2015Self}, without losing important information.

Video can be summarized at several levels of abstraction: keyframes~\cite{gong14diverse, lee2012discovering, liu2010hierarchical, mundur2006keyframe, zhang1997integrated}, segment or shot-based skims~\cite{laganiere2008video,lu2013story, nam2002event, ngo2003automatic}, story-boards~\cite{furini2010stimo, goldman2006schematic}, montages~\cite{kang2006space,sun2014salient} or video synopses~\cite{pritch2007webcam}. In this paper, we focus on developing learning algorithms for selecting keyframes or subshots from a video sequence. Namely, the input is a video and its subshots and the output is an (ordered) subset of the frames or subshots in the video. 

Inherently, summarization is a structured prediction problem where the decisions on whether to include or exclude certain frames or subshots into the subset are interdependent.  This is in sharp contrast to typical classification and recognition tasks  where the output is  a single label. 

The structured nature of subset selection presents a major challenge. Current approaches rely heavily on several heuristics to decide the desirability of each frame: representativeness~\cite{hong2009event, khosla2013large, ngo2003automatic}, diversity or uniformity~\cite{liu2002optimization, zhang1997integrated}, interestingness and relevance~\cite{kang2006space, lee2012discovering, lu2013story, ma2002user, ngo2003automatic}. However, combining those frame-based properties to output an optimal subset remains an open and understudied problem. In particular, researchers are hampered by the lack of knowledge on the ``global'' criteria human annotators presumably optimize when manually creating a summary. 

Recently, initial steps investigating \emph{supervised} learning for video summarization have been made.   They demonstrate  promising results~\cite{gong14diverse, Gygli2015video}, often  exceeding the conventional unsupervised clustering of frames. The main idea is to use a training set of videos and human-created summaries as targets to adapt the parameters of a subset selection model to optimize the quality of the summarization. If successful, a strong form of supervised learning would extract high-level semantics and cues from human-created summaries to guide summarization.

\begin{figure*}
\centering
\includegraphics[width=0.8\textwidth]{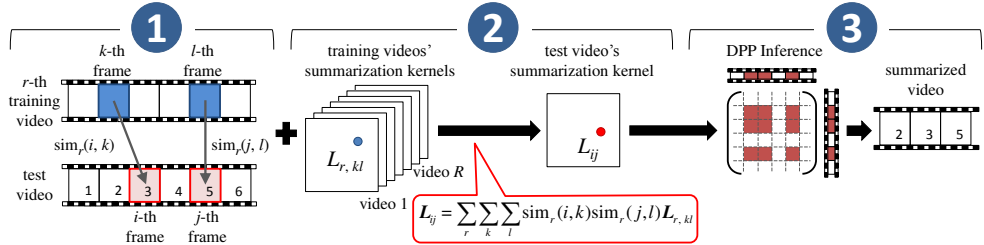}
\caption{The conceptual diagram of our approach, which leverages the intuition that \emph{similar videos share similar summary structures}. The main idea is nonparametric structure transfer, ie, transferring the subset structures in the human-created summaries (blue frames) of the training videos  to a new video. Concretely, for each new video, we first compute frame-level similarity between training and test videos (i.e., $\mathsf{sim}(\cdot, \cdot)$, cf. eq.~(\ref{eFrameSim})). Then, we encode the summary structures in the training videos with kernel matrices made of binarized pairwise similarity among their frames. We combine those structures, factoring the pairwise similarity between the training and the test videos, into a kernel matrix that encodes the summary structure for the test video, cf. eq.~(\ref{eSyn}).  Finally, the summary is decoded by inputting the kernel matrix to a  probabilistic model called the determinantal point process (DPP) to extract a globally optimal subset of frames.
}
\label{fConcept}
\vskip -1em
\end{figure*}
 
Supervised learning for structured prediction is a challenging problem in itself.  Existing parametric techniques typically require a complex model with sufficient annotated data to represent highly complicated decision regions in a combinatorially large output space. 
 In this paper,  we explore a \emph{nonparametric} supervised learning approach to summarization. Our method is motivated by the observation that \emph{similar videos share similar summary structures}. For instance,  suppose we have a collection of videos of wedding ceremonies inside churches. It is quite likely good summaries for those videos would all contain frames portraying brides proceeding to the altar, standing of the grooms and their best men, the priests' preaching, the exchange of rings, etc.  Thus, if one such  video is annotated with human-created summaries, a clever algorithm could essentially ``copy and paste'' the relative positions of the extracted frames in the annotated video sequence and apply them to an unannotated one to extract relevant frames.  Note that this type of transferring summary structures across videos need \emph{not} assume a  precise matching of visual appearance in corresponding frames --- there is no need to have the same priest as long as the frames of the priests in each video are sufficiently different from other frames to be ``singled out'' as possible candidates.

The main idea of our approach centers around this intuition, that is, \emph{non-parametric learning from exemplar videos} to transfer summary structures to novel input videos.   In recent years, non-parametric methods in the vision literature have shown great promise in letting the data ``speak for itself'', though thus far primarily for traditional categorization or regression tasks (e.g., label transfer for image recognition~\cite{liu2011nonparametric, torralba200880} or scene completion~\cite{hays2007scene}).  

How can summarization be treated non-parametrically?   A naive application of non-parametric learning to video summarization would treat keyframe selection as a binary classification problem---matching each frame in the unannotated test video to the nearest human-selected keyframes in some training video, and deciding independently per frame whether it should be included in or excluded from the summary.
Such an approach, however, conceptually fails on two fronts.  First, it fails to account for the relatedness \emph{between} a summary's keyframes.  Second, it limits the system to inputs having very similar frame-level matches in the annotated database, creating a data efficiency problem.  

Therefore, rather than transfer simple relevance labels, our key insight is to transfer the \emph{structures} implied by subset selection.  We show how kernel-based representations of a video's frames (subshots) can be used to detect and align the meta-cues present in selected subsets.  In this way, we compose novel summaries by borrowing recurring structures in exemplars for which we have seen both the source video and its human-created summary.  A conceptual diagram of our approach is shown in Fig.~\ref{fConcept}.

In short, our main contributions are an original modeling idea that leverages non-parametric learning for structured objects (namely, selecting subsets from video sequences), a summarization method that  advances the frontier of supervised learning  for video summarization, and an extensive empirical study validating the proposed method and attaining far better summarization results than competing methods on several benchmark datasets.

The rest of the paper is organized as follows. In section~\ref{sApproach}, we describe our approach of nonparametric structure transfer. We report and analyze experimental results in section~\ref{sExp} and conclude in section~\ref{sConclusion}.

%% file: related.tex
\section{Related Work}
\label{sRelated}

A variety of video summarization techniques have been developed in the literature~\cite{money2008video, truong2007video}.  Broadly speaking, most methods first compute visual features at the frame level, then apply some selection criteria to prioritize frames for inclusion in the output summary.

Keyframe-based methods select a subset of frames to form a summary, and typically use low-level features like optical flow~\cite{wolf1996key} or image differences~\cite{zhang1997integrated}.  Recent work also injects high-level information such as object tracks~\cite{liu2010hierarchical} or ``important" objects~\cite{lee2012discovering}, or takes user input to generate a storyboard~\cite{goldman2006schematic}.  In contrast, video skimming techniques first segment the input into subshots using shot boundary detection. The summary then consists of a selected set of representative subshots~\cite{laganiere2008video, nam2002event, ngo2003automatic}.

Selection criteria for summaries often aim to retain diverse and representative frames~\cite{hong2009event, khosla2013large, liu2002optimization, ngo2003automatic, zhang1997integrated}.  Another strategy is to predict object and event saliency~\cite{kang2006space,lee2012discovering, ma2002user,ngo2003automatic}, or to pose summarization as an anomaly detection problem~\cite{kim2009observe, zhao2014quasi}.  When the camera is known to be stationary, background subtraction and object tracking offer valuable cues about the salient entities in the video~\cite{feng2012online, pritch2007webcam}.

Whatever the above choices, existing methods are almost entirely \emph{unsupervised}.  For example, they employ clustering to identify groups of related frames, and/or manually define selection criteria based on intuition for the problem.  Some prior work includes supervised learning components (e.g., to generate regions with learned saliency metrics~\cite{lee2012discovering}, train classifiers for canonical viewpoints~\cite{khosla2013large}, or recognize fragments of a particular event category~\cite{potapov2014category}), but they do not learn the subset selection procedure itself.

Departing from unsupervised methods, limited recent work formulates video summarization as a subset selection problem  
\cite{gong14diverse, Gygli2015video, kim2014joint, Xu2015joint}. This enables supervised learning, exploiting knowledge encoded in human-created summaries. In~\cite{Gygli2015video}, a submodular function optimizes a global objective function of the desirability of selected frames, while \cite{gong14diverse} uses a probabilistic model that  maximizes the probability of the ground-truth subsets.

The novelty of our approach is to learn non-parametrically from exemplar training videos to transfer summary structures to test videos. In contrast to previous parametric models~\cite{gong14diverse, Gygli2015video}, non-parametric learning  generalizes to new videos by directly exploiting patterns in the training data. This has the advantage of generalizing locally within highly nonsmooth regions:   as long as a test video's ``neighborhood'' contains an annotated training video, the summary structure of that training video will be transferred. In contrast, parametric techniques typically require a complex model with sufficient annotated data to parametrically represent those regions.  Our non-parametric approach also puts design power into flexible kernel functions, as opposed to relying strictly on combinations of hand-crafted criteria (e.g., frame interestingness, diversity, etc.).

%% file: approach.tex
\section{Approach}
\label{sApproach}

We cast the process of extracting a summary from a video as selecting a subset of items (i.e., video frames) from a ground set (i.e., the whole video). Given a corpus of videos and their human-created summaries, our learning algorithm learns the optimal criteria for subset selection and applies them to unseen videos to extract summaries.

The first step is to decide on a subset selection model that can output a structure (i.e., an ordered subset). For such structured prediction problems, we focus on the \emph{determinantal point process} (DPP)~\cite{kulesza2012determinantal} which has the benefits of being more computationally tractable than many probabilistic graphical models~\cite{koller2009probabilistic}. Empirically, DPP  has been successfully applied to documentation summarization~\cite{kulesza2011learning}, image retrieval~\cite{gillenwater2012discovering} and more recently, to video summarization~\cite{chao2014large, gong14diverse}.

We will describe first  DPP and how it can be used for video summarization. We then describe our main approach in detail, as well as its several extensions.

\subsection{Background}
\label{sDPP}

Let $\ground=\{1,2,\cdots,\cN\}$ denote a (ground) set of $\cN$ items, such as video frames. The ground set has  $2^{\cN}$ subsets. The DPP over the $\cN$ items assigns a probability to each of those subsets. Let $\vy \subseteq \ground$ denote a subset and the probability of selecting it is given by
\begin{equation}
P(\vy; \mL) = \frac{\det(\mL_{\vy})}{\det(\mL+\mI)},
\end{equation}
where  $\mL$ is a symmetric, positive semidefinite matrix and $\mI$ is an identity matrix of the same size of $\mL$. $\mL_{\vy}$ is the principal minor (sub-matrix) with rows and columns from $\mL$ indexed by the integers in $\vy$.

DPP can be seen conceptually as a fully connected $\cN$-node Markov network where the nodes correspond to the items. This network's node-potentials are given by the diagonal elements of $\mL$ and the edge potentials are given by the off-diagonal elements in $\mL$. Note that those ``potentials'' cannot be arbitrarily assigned --- to ensure they form a valid probabilistic model, the matrix $\mL$ needs to be positive semidefinite. Due to this constraint, $\mL$ is often referred to as a kernel matrix whose elements can be interpreted as measuring the  pairwise compatibility.

Besides computational tractability which facilitates parameter estimation, DPP has an important modeling advantage over standard Markov networks.  Due to the celebrated Hammersley-Clifford Theorem, Markov networks cannot model  distributions where there are zero-probability events.  On the other hand, DPP is capable of assigning zero probability to absolutely impossible (or inadmissible) instantiations of random variables.  

To see its use for video summarization, suppose there are two frames that are identical.  For keyframe-based summarization, any subset containing such identical frames should be ruled out by being assigned zero probability. This is impossible  in Markov networks --- no matter how small, Markov networks will assign strictly positive probabilities to an exponentially large number of subsets containing identical frames. For DPP, since the two items are identical, they lead to two identical columns/rows in the matrix $\mL$, resulting a determinant zero (thus zero probability) for those subsets. Thus, DPP naturally encourages selected items in the subset to be diverse, an important objective for summarization and information retrieval~\cite{kulesza2012determinantal}. 

The mode of the distribution is the most probable subset
\begin{equation}
\vy^* = \argmax_{\vy} P(\vy;{\mL}) = \argmax_{\vy} \det (\mL_{\vy}).
\label{eMAP}
\end{equation}
This is an NP-hard combinatorial optimization problem, and there are several approaches to obtaining approximate solutions~\cite{gillenwater2012near, kulesza2012determinantal}.

The most crucial component in a DPP is its kernel matrix $\mL$. To apply DPP to video summarization, we define the ground set as the  frames  in a video and identify the most desired summarization as the MAP inference result of eq.~(\ref{eMAP}). We compute $\mL$ with a bivariate function over two frames --- we dub it the \emph{summarization kernel}:
\begin{equation}
L_{ij} = \vphi(\vv_i)\T\vphi(\vv_j)
\end{equation}
where $\vphi(\cdot)$ is a function of the features $\vv_i$ (or $\vv_j$) computed on the $i$-th (or the $j$-th ) frames. There are several choices. For instance, $\vphi(\cdot)$ could be an identity function, a nonlinear mapping implied by a Gaussian RBF kernel, or the output of a neural network~\cite{gong14diverse}.

As each different video needs to have a different kernel, $\vphi(\cdot)$ needs to be identified from a sufficiently rich function space so it generalizes from modeling the training videos to new ones. If the videos are substantially different, this generalization can be  challenging, especially when there are not enough annotated videos with human-created summaries. Our approach overcomes this challenge by directly using the training videos' kernel matrices, as described below.

\subsection{Non-parametric video summarization}
\label{sNonDPP}

Our approach differs significantly from existing summary methods, including those based on DPPs.  Rather than learn a single function $\vphi(\cdot)$ and discard the training dataset, we construct $\mL$ for every unannotated video by comparing it to the annotated training videos. This construction exploits two sources of information: 1) how similar the new video is to annotated ones, and 2) how the training videos are summarized. The former can be inferred directly by comparing visual features at each frame, while the latter can be ``read off'' from the human-created training summaries.

We motivate our approach with an idealized example that provides useful insight.  Let us assume we are given a training set of videos and their summaries  $\mathcal{D}= \{(\ground_r, \vy_r)\}_{r=1}^{\cR}$ and a new video $\ground$ to be summarized. Suppose this new video is very similar --- we define similarity more precisely later --- to a particular $\ground_r$ in $\mathcal{D}$.  Then we can reasonably assume that the summary $\vy_r$ might work well for the new video. As a concrete example, consider the case where both $\ground$ and $\ground_r$ are videos for wedding ceremonies inside churches. We anticipate seeing similar events across both videos: brides proceeding to the altars, priests delivering speeches, exchanging rings etc.  Moreover, similarity  in their summaries exists on the higher-order \emph{structural} level: the relative positions of the summary frames of $\vy_r$ in the sequence $\ground$ are an informative prior on where the frames of the summary $\vy$ should be in the new video $\ground$.  Specifically, as long as we can link the test video to the training video by identifying  similar frames,\footnote{This task is itself not trivial, of course, but it does have the benefit of a rich literature on image matching and recognition work, including efficient search strategies.} we can ``copy down''---transfer---the positions of $\vy_r$ and lift the corresponding frames in $\ground$ to generate its summary $\vy$. 

While this intuition is conceptually analogous to the familiar paradigm of nearest-neighbor classification, our approach is significantly different. The foremost is that, as discussed in section~\ref{sIntro}, we cannot select frames independently (by nonparametrically learning its similarity to those in the training videos). We need to transfer summary structures which encode interdependencies of selecting frames. Therefore, a naive solution of representing videos with fix-length descriptors in Euclidean space and literally pretending their summaries are ``labels'' that can be transferred to new data is flawed.

The main steps of our approach are outlined in Fig.~\ref{fConcept}. We describe them in detail in the following.

\paragraph{Step 1: Frame-based visual similarity}  To infer similarity across videos, we experiment with common ones in the computer vision literature for calculating  frame-based similarity from visual features $\vv_i$ and $\vv_k$ extracted from the corresponding frames:
\begin{equation}
\begin{aligned}
\mathsf{sim}^1(i, k) & = \vv_i\T\vv_k\\
\mathsf{sim}^2(i, k) & = \exp\{-\twonorm{\vv_i - \vv_k}/\sigma\} \\
\mathsf{sim}^3(i, k) & = \exp\{-(\vv_i - \vv_k)\T\mOmega(\vv_i - \vv_k)\},
\end{aligned}
\label{eFrameSim}
\end{equation}
where $\sigma$ and $\mOmega$ are adjustable parameters (constrained to be positive or positive definite, respectively). These forms of similarity measures are often used in vision tasks and are quite flexible, e.g., one can learn the kernel parameters for $\mathsf{sim}^3$.  However, they are not the focus of our approach --- we expect more sophisticated ones will only benefit our learning algorithm. We also expect high-level features (such as interestingness, objectness, etc.) could also be beneficial.  In section~\ref{sExt} we discuss a generalization to replace frame-level similarity with subshot-level similarity.

\paragraph{Step 2: Summarization kernels for training videos} The summarization kernels $\{\mL_r\}_{r=1}^{\cR}$ are not given to us in the training data. However, note that the crucial function of those kernels is to ensure that when used to perform the MAP  inference in eq.~(\ref{eMAP})  to identify the summary on the training video $\ground_r$ , it will lead to the correct summarization $\vy_r$ (which is in the training set). This prompts us to define the following \emph{idealized summarization kernels}
\begin{equation}
\mL_r = \alpha_r
  \begin{bmatrix}
    \delta(1 \in \vy_r) & 0 & \cdots & 0  \\
    0 & \delta(2 \in \vy_r)& \ddots  & \vdots  \\
    \vdots & \ddots & \ddots & 0 \\
    0 & \cdots & 0 & \delta(\cN_r \in \vy_r) \\
    \end{bmatrix}
\end{equation}
or more compactly,
\begin{align}
\mL_r = \alpha_r \mathsf{diag}( \{\delta(n \in \vy_r)\}_{n=1}^{\cN_r}),
\end{align}
where $\mathsf{diag}$ turns a vector into a diagonal matrix, $\cN_r$ is the number of frames in $\ground_r$ and   $\alpha_r > 1$ is an adjustable parameter.  The structure of $\mL_r$ is intuitive: if a frame is in the summary $\vy_r$, then its corresponding diagonal element is $\alpha_r$, otherwise 0.  It is easy to verify that $\mL_r$ indeed gives rise to the correct summarization.  Note that $\alpha_r >1$ is required.
If $\alpha_r = 1$, any subset of $\vy_r$ is a solution to the MAP inference problem (and we will be getting a shorter summarization). If $\alpha_r <1$, the empty set would be the summary (as the determinant of an empty matrix is 1, by convention).

\paragraph{Step 3: Transfer summary structure} Our aim is now to transfer the structures encoded by the idealized summarization kernels from the training videos to a new (test) video $\ground$. To this end, we synthesize $\mL$ for new video $\ground$ out of $\{\mL_r\}$.

Let $i$ and $j$ index the video frames in $\ground$, with  $k$ and $l$ for a specific training video $\ground_r$.  Specifically, the ``contribution'' from $\ground_r$ to $\mL$ is given by
\begin{equation}
r_{ij} = \sum_{k} \sum_{l} \mathsf{sim}_r(i, k)\mathsf{sim}_r(j, l) L_{r, kl}
\label{eSyn}
\end{equation}
where $L_{r, kl}$ is the element in $\mL_r$, and $\mathsf{sim}_r(\cdot, \cdot)$ measures frame-based (visual) similarity between frames of $\ground$ and $\ground_r$.  

Fig.~\ref{fConcept} illustrates graphically how frame-based similarity enables transfer of structures in training summaries.  We gain further insights by examining the case when the frame-based similarity $\mathsf{sim}_r(\cdot, \cdot)$ is sharply peaked --- namely, there are very good matches between specific pairs of frames (an assumption likely satisfied in the running example of summarizing wedding ceremony videos)
\begin{equation}
\begin{aligned}
\mathsf{sim}_r(i, m) &\gg \mathsf{sim}_r(i, k), &\forall\ k \ne m\\
\mathsf{sim}_r(j, n) &\gg \mathsf{sim}_r(j, l), &\forall\ l \ne n.
\end{aligned}
\end{equation}
Under these conditions,
\begin{equation}
r_{ij} \approx  \mathsf{sim}_r(i, m)\mathsf{sim}_r(j, n) L_{r, mn}.
\end{equation}
Intuitively, if $\ground$ and $\ground_r$ are precisely the same video (and the video frames in $\ground_r$ are sufficiently visually different), then the  matrix $\mL$ would be very similar to $\mL_r$.  Consequently, the summarization $\vy_r$, computed from $\mL_r$, would be a good summary for $\ground$.

To include all the information in the training data, we sum up the contributions from all $\ground_r$ and arrive at
\begin{equation}
L_{ij} = \sum_r r_{ij}.
\end{equation}
We introduce a few shorthand notions.  Let $\mS_r$ be a $\cN \times \cN_r$ matrix whose elements are $\mathsf{sim}_r(i, k)$, the frame-based similarity between $\cN$ frames in $\ground$ and $\cN_r$ frames in $\ground_r$.  The kernel matrix $\mL$ is thus
\begin{equation}
\mL = \sum_r  \mS_r \mL_r \mS_r\T.
\label{eSummaryKernel}
\end{equation}
Note that, $\mL$ is for the test video with $\cN$ frames --- there is no need for all the videos have the same length.

\paragraph{Step 4: Extracting summary} Once we have computed  $\mL$ for the new video, we use the MAP inference eq.~(\ref{eMAP}) to extract the summary as the most probable subset of frames.

\subsection{Learning}

Our approach requires adjusting parameters such as $\valpha = \{\alpha_1, \alpha_2, \cdots, \alpha_\cR\}$ for the ideal summarization kernels and/or $\mOmega$ for computing frame-based visual similarity eq.~(\ref{eFrameSim}). We use maximum likelihood estimation to estimate those parameters. Specifically, for each video in the training dataset, we pretend it is a new video and formulate a kernel matrix
\begin{equation}
\hat{\mL}_q = \sum_r \mS_r^q \mL_r {\mS_r^q}\T, \forall, q = 1, 2, \cdots, \cR.
\end{equation}
We optimize the parameters such that the ground-truth summarization $\vy_q$ attains the highest probability under $\hat{\mL}_q$,
\begin{equation}
\valpha^* = \arg\max_{\valpha} \sum_{q=1}^\cR \log P(\vy_q; \hat{\mL}_q).
\label{eMLE}
\end{equation}
We can formulate a similar criterion to learn the $\mOmega$ parameter for  $\mathsf{sim}^3(\cdot, \cdot)$. We carry out the optimization by gradient descent.  In our experiments, we set $\sigma$ for $\mathsf{sim}^2$ to be 1, with features normalized to have unit norm. Additional details are in the Suppl. and omitted here for brevity.  

\subsection{Extensions}
\label{sExt}

\paragraph{Category-specific summary transfer}  Video datasets labeled with semantically consistent categories  have been emerging~\cite{potapov2014category,Song2015TVSum}. We view  categories as valuable prior information that can be exploited by our nonparametric learning algorithm. Intuitively, videos from the same category (activity type, genre, etc.) are likely to be similar in part, not only in visual appearance but also in high-level semantic cues (such as how key events are temporally organized), resulting in a similar summary structures. We extend our method to take advantage such optional side information in two ways:
\begin{itemize}
\item \textsf{hard} transfer.  We compare the new video from a category $c$ only to the training videos from the same category $c$. Mathematically, for each video category, we learn a category-specific set of $\valpha^{(c)} = \{\alpha_1^{(c)}, \alpha_2^{(c)}, \cdots, \alpha_\cR^{(c)}\}$ such that $\alpha_r^{(c)}>0$ only when the training video $r$ belongs to category $c$. 
\item \textsf{soft} transfer.  We relax the requirement in \textsf{hard} transfer such that $\alpha_r^{(c)}>0$  even if the $r$th training video is not from the category $c$. Note that while we utilize structural information from all training videos, the way we use them still depends on the test video's category.
\end{itemize}

\paragraph{Subshot-based summary transfer} Videos can also be summarized at the level of subshots. As opposed to selecting keyframes, subshots contain short but contiguous frames, giving a glimpse of a key event. We next extend our subset selection algorithm to select a subset of subshots.

To this end, in our conceptual diagram as in Fig.~\ref{fConcept}, we replace computing frame-level similarity with subshot-level similarity, where we compare subshots between the training videos and the new video. We explore two possible ways to compute the frame-set to frame-set similarity:
\begin{itemize}
\item Similarity between averaged features. We represent the subshots using the averaged frame-level feature vectors within each subshot. We then compute the similarity using the previously defined $\mathsf{sim}(\cdot, \cdot)$.
\item Maximum similarity.  We compute pairwise similarity between frames within the subshots and select the maximum value as the similarity between the subshots.
\end{itemize}
Both of these two approaches reduce the reliance on frame-based similarity defined in the global frame-based descriptors of visual appearance, loosening the required visual alignment for discovering a good match---especially with the latter max operator, in principle, since it can ignore many unmatchable frames in favor of a single strong link within the subshots.  Moreover, the first approach can significantly reduce the computational cost of nonparametric learning as the amount of pairwise-similarity computation now depends on  the number of subshots, which is substantially smaller than the number of frames.  

\subsection{Implementation and computation cost}

Computing $\mS_r$ in eq.~(\ref{eSummaryKernel}) is an $O(\cN \times \sum_r \cN_r)$ operation. 
For long videos, several approaches will reduce the cost significantly. First, it is a standard procedure to down-sample the video (by a factor of 5-30) to reduce the number of frames for keyframe-based summarization. Our subshot-based summarization can also reduce the computation cost, cf. section~\ref{sExt}. Generic techniques should also help --- $\mathsf{sim}(\cdot, \cdot)$ computes various forms of distances among visual feature vectors. Thus, many fast search techniques apply, such as locality sensitive hashing or tree structures for nearest neighbor searches.

%% file: experiment.tex
\section{Experiment}
\label{sExp}

We validate our approach on five benchmark datasets.  It  outperforms competing methods in many settings. We also analyze its strengths and weaknesses.

\begin{table*}
\centering
\small
\caption{Key characteristics of datasets used in our empirical studies. Most videos in these datasets have a duration from 1 to 5 minutes. }
\label{tDataset}
\vskip 0.25em
\begin{tabular}{c|c|c|c|c|c|c}
\centering{Dataset} & \shortstack{\# of \\video} &\shortstack{ \# of \\category}   & \shortstack{\# of Training\\ videos} & \shortstack{\# of Test\\ video} & \shortstack{Type of \\ summarization} & \shortstack{Evaluation metrics\\ F-score in matching}\tabularnewline
\hline 
Kodak & 18 & -   & 14 & 4 & keyframe & \multirow{2}{*}{selected frames}\tabularnewline
\cline{1-6} 
OVP & 50 & - & 40 & 10 & keyframe &\tabularnewline
\cline{1-7} 
Youtube & 31 & 2   & 31 & 8 & keyframe; subshot & selected frames; frames in selected subshots\tabularnewline
\hline
SumMe & 25 & -  & 20 & 5 & subshot & frames in selected subshots \tabularnewline
\cline{1-7} 
MED  & 160 & 10   & 128 & 32 & subshot & matching selected subshots \tabularnewline
\hline
\end{tabular}
\vskip -1em
\end{table*} 

\subsection{Setup}
\label{setup}
\paragraph{Data} For \emph{keyframe-based} summarization, we experiment on three video datasets: the \textbf{Open Video Project (OVP)} \cite{openvideo, de2011vsumm}, the \textbf{YouTube} dataset \cite{de2011vsumm}, and the \textbf{Kodak} consumer video dataset \cite{luo2009towards}. All the 3 datasets were used in \cite{gong14diverse} and we follow the procedure described there to preprocess the data, and to generate training ground-truths from multiple human-created summaries.  For the YouTube dataset, in the following, we report results on 31 videos after discarding 8 videos that are neither ``Sports'' nor ``News'' such that we can investigate category-specific video summarization (cf. sec.~\ref{sExt}). In Suppl., we report results on the original dataset.

For \emph{subshot-based} summarization (cf. sec.~\ref{sExt}), we experiment on three video datasets: the portion of \textbf{MED} with 160 annotated summaries ~\cite{potapov2014category}, \textbf{SumMe}~\cite{gygli2014creating} and \textbf{YouTube}. Videos in MED are pre-segmented into subshots with the Kernel Temporal Segmentation (KTS) method \cite{potapov2014category} and we observe those subshots. For SumMe and YouTube, we apply KTS to generate our own sets of subshots. MED has 10 well-defined video categories allowing us to experiment with category-specific video summarization on it too. SumMe does not have semantic category meta-data. Instead, its video contents have a large variation in visual appearance and can be classified according to shooting style: still camera, egocentric, or moving. 
Table~\ref{tDataset} summarizes key characteristics of those datasets with details in the Suppl.

\paragraph{Features} For OVP/YouTube/Kodak/SumMe, we encode each frame with an $\ell^2$-normalized 8192-dimensional Fisher vector \cite{perronnin2007fisher}, computed from SIFT features. For OVP/YouTube/Kodak, we also use color histograms. We also experimented with features from a pre-trained convolutional nets (CNN), details in the Suppl. For MED, we use the provided 16512-dimensional Fisher vectors. 

\paragraph{Evaluation metrics}
As in ~\cite{gong14diverse, gygli2014creating, Gygli2015video} and other prior work, we evaluate automatic summarization results (A) by comparing them to the human-created summaries (B) and reporting the standard metrics of F-score (F), precision (P), and recall (R) --- their definitions are given in the Suppl. 

For OVP/YouTube/Kodak, we follow~\cite{gong14diverse} and utilize the VSUMM package \cite{de2011vsumm} for finding matching pairs of frames.  
For SumMe, we follow the procedure in ~\cite{gygli2014creating, Gygli2015video}. More details are in the Suppl.

\paragraph{Implementation details} For each dataset, we randomly choose 80\% of the videos for training and use the remaining 20\% for testing, repeating for 5 or 100 rounds so that we can calculate averaged performance and standard errors. 
To report existing methods' results, we use prior published numbers when possible.  We also implement the  \textsc{vsumm} approach~\cite{de2011vsumm} and obtained code from the authors for \textsf{seqDPP}~\cite{gong14diverse} in order to apply them to several datasets.  We follow the practices in \cite{gong14diverse} so that we can summarize videos sequentially. For MED, we implement KVS ~\cite{potapov2014category} ourselves.  

\subsection{Main Results}

We summarize our key findings in this section. For more details, please refer to the Suppl.

In Table~\ref{tOYKSimple}, we compare our approach to both supervised and unsupervised methods for video summarization. We report published results in the table as well as results from our own implementation of  some methods. Only the best variants of all methods are quoted and presented; others are deferred to the Suppl. 

On all but one of the five datasets (OVP), our nonparametric learning method achieves the best results.  In general,  the supervised methods  achieve better results than the unsupervised ones.  Note that even for datasets with a variety of videos that are not closely visually similar (such as SumMe), our approach attains the best result---it indicates our method of transferring summary structures is effective, able to build on top of even crude frame similarities.

\begin{table*}[t]
\centering
\small
\caption{Performance (F-score) of various video summarization methods.  Numbers followed by citations are from published results. Others are from our own implementation.  Dashes denote unavailable/inapplicable dataset-method combinations.  }
\vspace{3pt}
\label{tOYKSimple}
\begin{tabular}{c|c|c|c|c|c|c|c|c|c|c}
& \multicolumn{6}{c}{Unsupervised} & \multicolumn{4}{|c}{Supervised}\\ \cline{2-11}
& \textsc{vsumm}$_1$& \textsc{vsumm}$_2$& DT& STIMO & KVS & Video MMR & SumMe  & Submodular  & \textsf{seqDPP}&  {\textbf{Ours}}\\
& \cite{de2011vsumm} & \cite{de2011vsumm}& \cite{mundur2006keyframe}& \cite{furini2010stimo} & \cite{potapov2014category} & \cite{li2010multi}  & \cite{gygli2014creating}  & \cite{Gygli2015video}  & \cite{gong14diverse}&  \\ \hline
Kodak & 69.5 & 67.6 & - & - & -  & -  & -  & -  &  78.9& \textbf{82.3}\\ \hline
OVP & 70.3 & 69.5& 57.6 & 63.4 &  -  &  -  & -  &-  & \textbf{77.7} & 76.5\\
\hline
YouTube & 58.7 &  59.9  & -  & -  & -  & -  & - & -  & 60.8 &\textbf{61.8} \\ \hline
MED & 28.9 & 28.8 & - & -  & 20.6 & - & - & - & -  & \textbf{30.7} \\ \hline
SumMe & 32.8 & 33.7 & - & - & - & 26.6 & 39.3 & 39.7 & - & \textbf{40.9} \\ \hline
\end{tabular}
\vskip -1em
\end{table*}

\subsection{Detailed analysis}

\paragraph{Advantage of nonparametric learning}  Nonparametric learning enjoys the property of generalizing locally. That is, as long as a test video has enough correctly annotated training videos in its neighborhood, the summary structures of those videos will transfer. A parametric learning method, on the other end,  needs to learn both the locations of those local regions and how to generalize within local regions.  If there are not enough annotated training videos covering the whole range of data space, it could be difficult for a parametric learning method to learn well. 

We design a simple example to illustrate this point. As it is difficult to assess ``similarity'' to derive nearest neighbors for video, we use a video's category to delineate those ``semantically near''. Specifically, we split YouTube's 31 videos into two piles, according to their categories ``Sports'' or ``News''.  We then construct seqDPP, a parametric learning model~\cite{gong14diverse}, using all the 31 videos, as well as ``Sports'' or ``News'' videos only to summarize test videos from either category. We then contrast to our method in the same setting.  Table~\ref{tNonparam} displays the results.

The results on the ``News'' category convincingly suggest that the nonparametric approach like ours can leverage the semantically-close videos to outperform the parametric approach with the same amount of annotated data---or even more data. (Note that, the difference on the ``Sports'' category is nearly identical.)
\begin{table}
\centering
\small
\caption{Advantage of nonparametric summary transfer}
\label{tNonparam}
\begin{tabular}{c|c|c||c|c}
Type of & \multicolumn{2}{c||}{seqDPP} & \multicolumn{2}{c}{Ours}\\ \cline{2-5}
 test video & all & same as test & all & same as test\\ \hline
 Sports &52.8 & 54.5 & 53.5  & 54.4 \\ \hline
 News & 67.9 & 67.7 &  66.9 & 69.1\\ \hline
 \end{tabular}
 \vskip -0.5em
 \end{table}
 
\paragraph{Advantage of exploiting category prior}  Table~\ref{tNonparam} already alludes to the fact that exploiting category side-information can improve summarization (cf.~contrasting the column of ``same as test'' to ``all'').  Now we investigate this advantage in  more detail. Table~\ref{tHardSoft} shows how our nonparametric summary transfer can exploit video category information, using the method explained in section~\ref{sExt}.

Particularly interesting are our results on the SumMe dataset, which itself does not provide semantically meaningful categories.  Instead, we generate two ``fake'' categories for it. We first collapse the 10 video categories in the dataset TVSum\footnote{We choose this one as it has raw videos for us to extract features and have a larger number of labeled videos for us to build a category classifier}~\cite{Song2015TVSum} into two super-categories (details in Suppl.) --- these two super-categories are semantically similar within each other, though they do not have obvious visual similarity to videos in the SumMe.   

We then build a binary classifier trained on TVSum videos but classify the videos in SumMe as ``super-category I'' and ``super-category II'' and then proceed as if they are ground-truth categories, as in MED and YouTube. Despite this independently developed dichotomy, the results on SumMe improve over using all video data together.

\begin{table}
\centering
\small
\caption{Video category information helps summarization}
\label{tHardSoft}
\vskip 0.25em
\begin{tabular}{c|c|c|c}
{Setting} & YouTube & MED & SumMe\\ \hline
w/o  category & 60.0 & 28.9 & 39.2 \\ \hline
category-specific \scriptsize{ $\mathsf{hard}$} & 61.5 & 30.4 & 40.9 \\ \hline
category-specific \scriptsize{ $\mathsf{soft}$} & 60.6 & 30.7 & 40.2 \\ \hline
\end{tabular}
\vskip -0.5em
\end{table}

\paragraph{Subshot-based summarization} In section~\ref{sExt}, we discuss an extension to summarize video at the level of selecting subshots. This extension not only reduces computational cost (as the number of subshots is significantly smaller than that of frames), but also provides additional means of measuring similarity across videos beyond frame-level visual similarity inferred from global frame-based descriptors.  Next we examine how such flexibility can ultimately improve keyframe-based summarization. Concretely,  we first perform subshot summarization, then pick the middle frame in each selected subshot as the output  keyframes. This allows us to directly compare to keyframe-based summarization using the same F-score metric.

Table~\ref{tSubshot} shows the results.  Subshot-based summarization clearly improves frame-based --- this is very likely due to the more robust similarity measures now computed at the subshot-level.   The improvement is more pronounced when a category prior is not used. One possible explanation is that measuring similarity on videos from the same categories is easier and more robust, whereas across categories it is noisier. Thus, when a category prior is not present, the subshot-based similarity measure benefits summarization more. 

\begin{table}
\centering
\small
\caption{Subshot-based summarization on YouTube}
\label{tSubshot}
\vspace{0.25em}
\begin{tabular}{c|c|c|c}
Category  &  Frame- & \multicolumn{2}{c}{Subshot-based}\\ \cline{3-4}
-specific & based & Mean-Feature &  Max-similarity \\ \hline
No& 60.0 & 60.7 &  60.9 \\ \hline
Yes   & 61.5 & 61.6 & 61.8\\ \hline
\end{tabular}
\vskip -1em
\end{table}

\vspace*{-0.15in}
\paragraph{Other detailed analysis in Suppl.} We summarize other analysis results as follows. We show how to improve frame-level similarity by learning better metrics. We also show deep features, powerful for visual category recognition, is not particularly advantageous comparing to shallow features. We also show how category prior can still be exploited even we do not know the true category of test videos.

\subsection{Qualitative analysis}

\begin{figure}
\centering
\includegraphics[width=0.99\columnwidth]{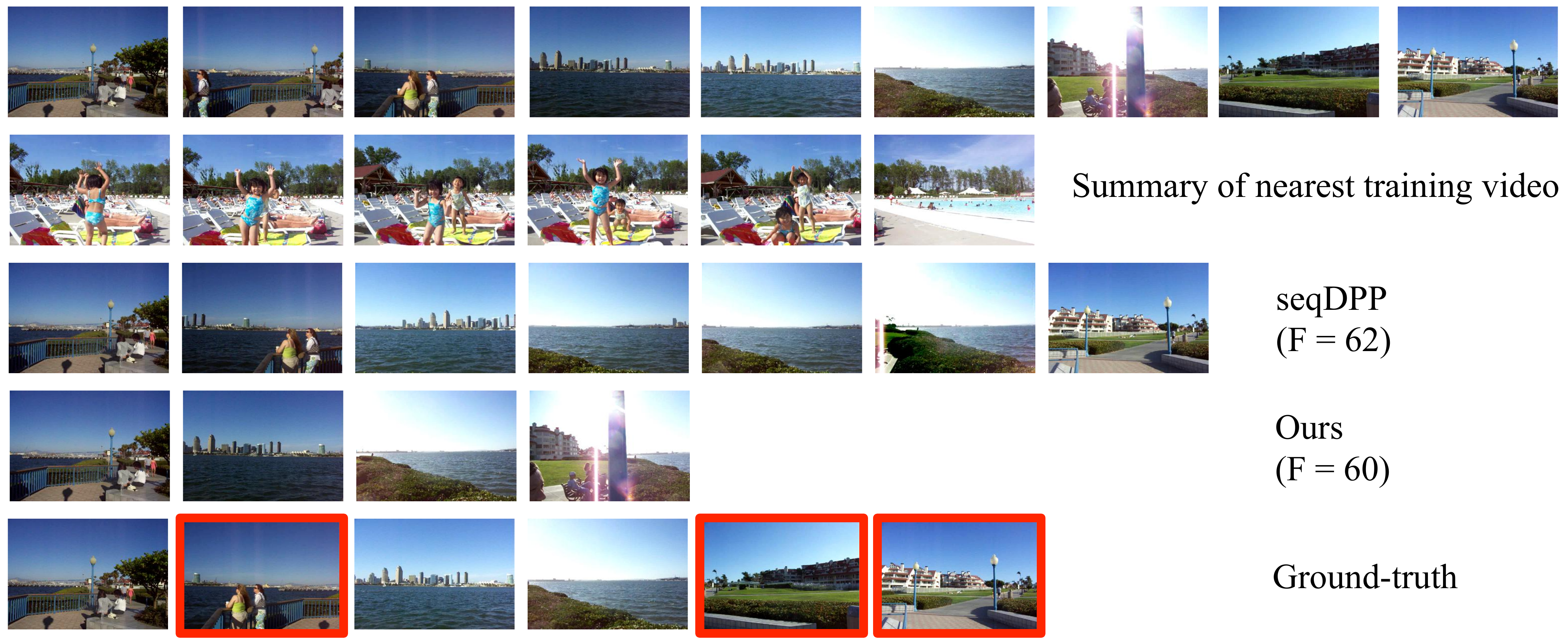}
\caption{A failure case by our approach. Our summary misses the last two frames from the ground-truth (red-boxed) as the test video (nature scene)  transfers from the nearest video with a semantically different category (beach activity).  See text.} 
\label{fFail}
\vskip -1em
\end{figure}

Fig.~\ref{fFail} shows a failure case by our method.
Here the test video depicts a natural scene, while its closest training video depicts beach activities. There is a visual similarity (e.g., in the swath of sky). However, semantically, these two videos do not seem to be relevant and it is likely difficult for the transfer to occur. In particular, our results miss the last two frames where there are a lot of grass. This suggests one weakness in our approach:  the formulation of our summarization kernel for the test video does not directly consider the relationship between its own frames --- instead, they interact through training videos.  Thus, one possible direction to avoid unreliable neighbors in the training videos is to rely on the visual property of the test video itself. 
This suggests future work on a hybrid approach with parametric and nonparametric aspects that complement each other.

Please see the Suppl. for more qualitative analysis and example output summaries.

%% file: conclusion.tex
\section{Conclusion}
\label{sConclusion}

We propose a novel supervised learning technique for video summarization. The main idea is to learn nonparametrically to transfer summary structures from training videos to test ones. We also show how to exploit side (semantic) information such as video categories and propose an extension for subshot-based summarization.  Our method achieves promising results on several benchmark datasets, compared to an array of nine existing techniques.

%% file: gradient.tex
\section{Learning}
\label{sMLE}

\subsection{Maximum Likelihood Estimation}
We use gradient descent to maximize the log-likelihood $ \sum_{q=1}^\cR \log P(\vy_q; \hat{\mL}_q)$, defined in (13) of the main text, w.r.t. $\valpha$ and $\mOmega$:
\begin{equation}
\label{Gradients1}
\begin{split}
(\valpha^*, \mOmega^*) = \arg\max_{\valpha, \mOmega} \sum_{q=1}^\cR \log P(\vy_q; \hat{\mL}_q).
\end{split}
\end{equation}
For brevity, we ignore the subscript $q$ in the following.
The corresponding partial derivatives w.r.t. $\valpha$ and $\mOmega$ are shown below, according to the chain rule:
\begin{align}
\label{Gradients1_0}
&\frac{\partial \log P(\vy; \hat{\mL})}{\partial \alpha_r} \nonumber\\
& = \sum_{m,n}\sum_{k,l}\frac{\partial \log P(\vy; \hat{\mL}) }{\partial (\hat{\mL})_{mn}}\frac{\partial  (\hat{\mL})_{mn}}{\partial \mL_{r, kl}}\frac{\partial  \mL_{r, kl}}{\partial \alpha_r}\\
&\nonumber\\	&\nonumber\\	
&\frac{\partial \log P(\vy; \hat{\mL}) }{\partial \mOmega}\nonumber\\
&= \sum_{r=1}^\cR  \sum_{m,n}\sum_{k,l}\frac{\partial \log P(\vy; \hat{\mL}) }{\partial (\hat{\mL})_{mn}}\frac{\partial (\hat{\mL})_{mn}}{\partial \mS_{r, kl}}\frac{\partial \mS_{r, kl}}{\partial \mOmega}\\\nonumber
\end{align}
To ease the computation in what follows, we define $\mM$ as a matrix with the same size of $\hat{\mL}$, and $\mM_{\vy} = (\hat{\mL}_{\vy})^{-1}$. All the other entries of $\mM$ are zero.

\subsection{Gradients with respect to the scale parameters}
\begin{equation}
\label{Gradients1_1}
\begin{split}
&\frac{\partial \log P(\vy; \hat{\mL})}{\partial \alpha_r} \\
&= \sum_{m,n}\sum_{k,l}\frac{\partial \log P(\vy; \hat{\mL}) }{\partial (\hat{\mL})_{mn}}\frac{\partial  (\hat{\mL})_{mn}}{\partial \mL_{r, kl}}\frac{\partial  \mL_{r, kl}}{\partial \alpha_r}\\
&= \textbf{1}^\top\{\Big( {\mS_r}^{\top}(\mM - (\hat{\mL}+ I)^{-1})\mS_r \Big)\circ \mI_r\}\textbf{1},\\
\end{split}
\end{equation}
where $\circ$ is the element-wise product. $\mI_r$ is a diagonal matrix (of the same size as $\mL_r$), with entries indexed by $\vy_r$ as one; otherwise, zero. $\textbf{1}$ is the all one vector with a suitable size.

\subsection{Gradients with respect to the transformation matrix}
\begin{equation}
\label{Gradients2_1}	
\begin{split}
&\frac{\partial \log P(\vy; \hat{\mL}) }{\partial \mOmega}\\
&= \sum_{r=1}^\cR  \sum_{m,n}\sum_{k,l}\frac{\partial \log P(\vy; \hat{\mL}) }{\partial (\hat{\mL})_{mn}}\frac{\partial (\hat{\mL})_{mn}}{\partial \mS_{r, kl}}\frac{\partial \mS_{r, kl}}{\partial \mOmega}\\
&= 4\mOmega\big( \Phi{\mC_r}{\Phi_{r}}^{\top}+\Phi_{r}\mC_r^{\top}{\Phi}^{\top}\\
& - (\Phi\mC_r^{(1)}{\Phi}^{\top} + \Phi_{r}\mC_r^{(2)}{\Phi_{r}}^{\top})\big),\\
\end{split}
\end{equation}
where $\mC_r, \mC_r^{(1)}, \mC_r^{(2)}$ are defined as:
\begin{equation}
\label{Gradients2_1_1}
\begin{split}
&\mC_r = ((\mM-(\hat{\mL}+\mI)^{-1})\mS_r\mL_{r})\circ \mS_r,\\
&\mC_r^{(1)} = \mathsf{diag}(\mC_r\textbf{1})\\
&\mC_r^{(2)} = \mathsf{diag}(\textbf{1}^\top\mC_r).\\
\end{split}
\end{equation}
$\Phi_r$ and $\Phi$ are the column-wise concatenation (matrices) of $\{\vphi_r\}$ and $\{\vphi\}$, respectively.

\subsection{Extension for sequential modeling}
\label{sSeqMode} 
Prior work seqDPP~\cite{gong14diverse} shows that the vanilla DPP is inadequate in capturing the  sequential nature of video data. In particular, DPP models a bag of items such that permutation will not affect selected subsets. Clearly, video frames cannot be randomly permuted without losing the semantic meaning of the video. The authors propose the sequential DPP (seqDPP), which sequentially extracts summaries from video sequences.

Extending our approach to this type of sequential modeling is straightforward due to its non-parametric nature. Briefly, we segment each video in the training dataset into smaller chunks and treat each chunk as a separate ``training'' video containing a subset of the original summary. We follow the same procedure as described above to formulate the base summarization kernels for those shorter videos and their associated summaries. Likewise, we learn the parameters $\valpha$ and $\mOmega$ for each segment.

During testing time when we need to summarize a new video $\ground$, we  segment the video $\ground$ in temporal order: $\ground = \ground_1 \cup \ground_2 \cdots \ground_T$. We then perform the sequential summarization procedure in the seqDPP method. Specifically, at time $t$, we formulate the ground set as $\mU_t = \ground_t \cup \vy_{t-1}$, i.e., the union of the video segment $t$ and the \emph{selected subset} in the immediate past. For the ground set $\mU_t$, we calculate its kernel matrix using the (segmented) training videos and their summaries. We then perform the extraction. The recursive extraction process is exactly the same as in \cite{gong14diverse} and we hence omit it here.

In our experiments, we use sequential modeling for the Kodak, OVP, YouTube and SumMe datasets. For the MED dataset and subshot-based experiments on YouTube dataset, we directly perform our model on each video without sequential modeling as the number of subshots for each dataset is few. We perform KTS~\cite{potapov2014category} to divide a video into segments.

%% file: datasets.tex
\section{Datasets}
\label{sData}
We validate our approach on five benchmark datasets. In this section we give detailed descriptions for each of them.
\label{sDat}
\subsection{Kodak, OVP and YouTube dataset}
\label{sDatKOY}
We perform keyframe-based video summarization on three video datasets: the Open Video Project (OVP) \cite{openvideo, de2011vsumm}, the YouTube dataset \cite{de2011vsumm}, and the Kodak consumer video dataset \cite{luo2009towards}. They contain 50, 39, and 18 videos, respectively. In the main text, to investigate the effectiveness of category-specific prior, we use Category Sports (16 videos) and News (15 videos), i.e., 31 out of 39 videos.
The OVP and YouTube datasets have five human-created frame-level summaries per video, while Kodak has one per video. Thus, for the first two datasets, we follow the procedure described in \cite{gong14diverse} to create an oracle summary per video. The oracle summaries are then used to optimize model parameters of various methods.

As suggested in \cite{de2011vsumm}, we pre-process the video frames as follows. We uniformly sample one frame per second for OVP and YouTube, two frames per second for Kodak (as the videos are shorter) to create the ground set $\ground$ for each video. We then prune away obviously uninformative frames, such as transition frames close to shot boundaries and near-monotone frames. On average, the resulting ground set per video contains 84, 128, and 50 frames for OVP, YouTube, and Kodak, respectively. We replace each frame in the human-annotated summary with the temporally closest one from the ground set, if that is not already contained in the ground set.

\subsection{SumMe dataset}

The SumMe dataset \cite{gygli2014creating, Gygli2015video} consists of 25 videos, with the average length being 2m40s. Similar to \cite{potapov2014category}, each video is cut and summarized by 15 to 18 people, and the average length of the ground-truth (shot-based) summary is 13.1\% of the original video. We thus perform subshot-based summarization on this dataset. 

The video contents in  SumMe dataset is heterogeneous and it  does not come with pre-defined categories.  However,  some of them  have various degrees of relatedness. For example, videos with title 'Kids playing in leaves' and 'Playing on water slide' are about children playing. Thus, it is interesting to investigate if category prior can help improve summarization on this dataset. 

Instead of manually labeling each video, we propose two synthetic categories and split videos into two partitions. We first collapse the 10 video categories in the dataset TVSum\footnote{We choose this one as it has raw videos for us to extract features and have a larger number of labeled videos for us to build a category classifier.}~\cite{Song2015TVSum} into two super-categories, denoted as Super-category I and II, respectively. Super-category I includes activities such as attempting bike tricks, flash mob gathering, park tour and parade, which mostly have raw videos with crowds, and Super-category II includes the rest types of activities  with much less people interaction. 
These two super-categories are semantically similar within each other, though they do not have obvious visual similarity to videos in the SumMe.
We then build a binary classifier trained on TVSum videos but classify the
videos in SumMe as Super-category I and Super-category II, and then proceed as if they are ground-truth categories, as in MED and YouTube.

Since the shot boundaries given by users vary even for the same video, to create an oracle summary of each video for training, we first score each \emph{frame} by the number of user summaries containing it. We then obtain our shot boundaries based on KTS~\cite{potapov2014category} and compute the shot-level scores by averaging the frame scores within each shot. Finally we rank the shots and select the top ones that are combined to have around 15\% of the original video to be the oracle summary for each video.

\subsection{MED dataset}
\label{MED_data}
The MED dataset, developed in \cite{potapov2014category}, is a subset of the TRECVid 2011 MED dataset. The  dataset has 12,249 videos: 2,389 videos in the following 10 categories (namely Birthday party, Changing a vehicle tire, Flash mob gathering, Getting a vehicle unstuck, Grooming an animal, Making a sandwich, Parade, Parkour, Repairing an appliance, and Working on a sewing project) and 9,860 from none of those (labeled as 'null'). 160 videos from the above 2,389 videos are annotated with summaries. For each video in the dataset, an annotator is asked to cut and score all the shots in a video (from 0 to 3). We thus perform subshot-based summarization on this dataset.

There are several constraints in using this dataset: the dataset does not provide the original videos, only boundaries of previously segmented shots and pre-computed features (on average 27 shots per video and their Fisher vectors). Those shot boundaries are often different from the shots determined by human annotators. 

To use this dataset for subshot-based summarization, we first need to combine all the human annotators' shot scores into oracle summaries for training. To this end, we first map each users' annotations to the pre-determined shots --- in particular, each frame in its pre-determined shot ``inherents'' the importance scores from the human annotators' scores of the subshots if the subshots contain the frame. We  then compute shot-level importance scores by taking an average of the frame importances with in each pre-determined subshot. We then select shots having the top $K$\% importance scores as ground-truth. We then use this ground-truth for training. In this paper, we test our model under different summary length, i.e. $K$ = 15 and 30.

\subsection{Features}

For the Kodak, OVP, and YouTube datasets, we extract Fisher vectors, color histograms, and CNN features (based on GoogLeNet \cite{szegedy2014going}). In our experiments, we use Fisher vectors and color histogram for these three dataset, except the comparison in section~\ref{shallow_deep}.
For the MED dataset, as the author didn't provide the raw video, we use the Fisher vectors given in the dataset.
For SumMe dataset, we use the output of layer-6 of AlexNet \cite{krizhevsky2012imagenet} as features. 

%% file: eval_metric.tex
\section{Evaluation protocols}
\label{sEval}
In all our experiments, we evaluate automatic summarization results (A) by comparing them to the human-created summaries (B) and reporting the F-score (F), precision (P), and recall (R), shown as follows:

\begin{align}
& \text{Precision} =\frac{\small{\#}\text{\small{matched pairs}}}{\#\text{\small{frames (shots) in A}}}\times 100\%,\\
& \text{Recall} =\frac{\#\text{\small{matched pairs}}}{\#\text{\small{frames (shots) in B}}}\times 100\%, \\
& \text{F-score} =\frac{P\cdot R}{0.5(P+R)}\times 100\%.
\end{align}

For datasets with multiple human-created summaries, we average or take the maximum over the number of human-created summaries to obtain the evaluation metrics for each video. We then average over the number of videos to obtain the metrics for the datasets. 

\subsection{Kodak, OVP and YouTube datasets}
We utilize the VSUMM package \cite{de2011vsumm} to evaluate video summarization results. Given two sets of summaries, one formed by an algorithm and the other by human annotators, this package outputs the maximum number of matched pairs of frames between them. Two frames are qualified to be a matched pair if their visual difference\footnote{VSUMM computes the visual difference automatically based on the color histograms.} is below a certain threshold, while each frame of one summary can be matched to at most one frame of the other summary. We then can measure F-score based on these matched pairs. 

\subsection{SumMe}
We follow the protocol in ~\cite{gygli2014creating, Gygli2015video}. As they constrain the summary length to $\leq$ 15\% of the video duration, we take the shots selected by our model (and if their duration exceeds 15\%, we further rank them by their values on the diagonal of $\mL$-kernel matrix --- higher values imply more important items, in our probabilistic framework for subset selection). We then take the highest ranked shots and stop when their total length exceeds 15\% of the video duration.

In computing the F-score, \cite{gygli2014creating, Gygli2015video} use the same formulation in comparing the summarization result to a user summary. However, when dealing with multiple user summaries on one video, instead of taking the average as in~\cite{gygli2014creating}, \cite{Gygli2015video} reports the highest F-score compared to all the users. In this paper, we follow~\cite{Gygli2015video}.

\subsection{MED}
\label{MED_eval}
For MED, we conduct subshot-based summarization.  We follow the same procedure to align each user's annotation with pre-determined shot boundaries in the dataset (cf. section 2.3 in this Suppl. Material) to create user summaries.  We then evaluate similarly as we have done on the Kodak, OVP and YouTube datasets, treating subshots as ``frames''. Since the KVS algorithm \cite{potapov2014category} cannot decide the summarization length but only scores each shot, we simply take the top $15$\% or $30$\% shots (ordered by scores) to be the summary of KVS.

%% file: supp_exp.tex
\section{Detailed experimental results and analysis}
\label{sRes}

\subsection{Benefits of learning visual similarity}
We further examine the effect of employing different measures of visual similarity. As shown in Table \ref{tLearnedW}, nonlinear similarity with the Gaussian-RBF kernel ($\mathsf{sim}^2$) generally outperforms linear similarity ($\mathsf{sim}^1$), while learning a non-identity metric $\mOmega$ ($\mathsf{sim}^3$)  improves only marginally the performance.
In the following experiments, we use $\mathsf{sim}^2$.

\begin{table}[t]
\centering
\caption{For our method, better frame-based visual similarities improve summarization quality. We report 100-round results for Kodak, OVP and 5-round results for the remaining.}
\vspace{10pt}
\label{tLearnedW}
\begin{tabular}{c|c|c|c}
& $\mathsf{sim}^1$ & $\mathsf{sim}^2$ & $\mathsf{sim}^3$\\ \hline
Kodak & 76.6\scriptsize{$\pm$0.6}   &80.2\scriptsize{$\pm$0.3}& \textbf{82.3}\scriptsize{$\pm$0.3}\\
OVP &  71.4\scriptsize{$\pm$0.3}  &75.3\scriptsize{$\pm$0.3} & \textbf{76.5}\scriptsize{$\pm$0.4} \\
\hline
YouTube\scriptsize{ $\mathsf{soft}$}  &  58.9\scriptsize{$\pm$1.7} & {60.6} \scriptsize{$\pm$1.4}& \textbf{60.9}\scriptsize{$\pm$1.3}\\
YouTube\scriptsize{ $\mathsf{hard}$}  &  59.6\scriptsize{$\pm$1.4} & \textbf{61.5}\scriptsize{$\pm$1.5} & \textbf{61.5}\scriptsize{$\pm$1.5}\\
\hline
\end{tabular}
\vskip 0.5em
\end{table}

\subsection{Results on the complete YouTube dataset}
The original YouTube dataset~\cite{de2011vsumm}, after excluding the cartoon videos as in~\cite{gong14diverse}, contains 39 videos with 8 of them in neither Sports nor News category. In the main text, we have used 31 videos mainly for the purpose to compare to summarization results exploiting category prior.

For maximum comparability to prior work on this dataset, we provide our summarization results on the 39 videos and compare to previous published results, as in Table \ref{tYoutube39}. 

\begin{table}
\centering
\small
\caption{Results on YouTube dataset (39 videos), all based on 100 rounds.}
\vspace{1em}
\label{tYoutube39}
\small
\begin{tabular}{c|c|c|c}
& \textsc{vsumm}$_1$ \cite{de2011vsumm}  & seqDPP \cite{gong14diverse} & \textbf{Ours} \\\hline
YouTube & 56.9\scriptsize{$\pm$0.5} & 60.3\scriptsize{$\pm$0.5} & 60.2\scriptsize{$\pm$0.7}\\
\hline
\end{tabular}
\end{table}

\subsection{Detailed results with category prior}

The results for the YouTube, SumMe and MED datasets are shown in Table~\ref{tCategoryYouTube}, \ref{tCategorySumMe} and \ref{tCategoryMED}, respectively.

We can see that both the \textsf{soft} category-specific and the \textsf{hard} category-specific outperform the no category-specific setting in most cases.  For example, a 'birthday party' video is likely to share structures with an 'outdoor activity' video, thus helping to summarize each other.

Additionally, even when the ground-truth categories for the testing videos are unknown and need to be determined with a category classifier, the category prior can still help.

\begin{table*}
\centering
\caption{\small Exploiting category prior with YouTube dataset, which contains totally 31 videos in Sport and News categories. We report F-score in each category as well as on the whole dataset, averaging over 5 rounds of experiments.} 
\small
\vspace{1em}
\label{tCategoryYouTube}
\begin{tabular}{c|c||c|c}
Setting & Testing video's category &  \multicolumn{2}{c}{Category prior not used} \tabularnewline
\hline 
\multirow{3}{*}{YouTube} & Sports & \multicolumn{2}{c}{53.5\scriptsize{$\pm$1.5}}  \tabularnewline
\cline{2-4}   
 & News & \multicolumn{2}{c}{66.9\scriptsize{$\pm$1.2}} \tabularnewline
\cline{2-4} 
 & Combined & \multicolumn{2}{c}{60.0\scriptsize{$\pm$1.3}}\tabularnewline
\hline 
\multicolumn{2}{c||}{} & Ground-truth category is used & Category determined by a classifier  \tabularnewline
\hline 
\multirow{3}{*}{YouTube\scriptsize{ $\mathsf{soft}$}} & Sports & 53.4\scriptsize{$\pm$1.5} & 53.4\scriptsize{$\pm$1.5}\tabularnewline
\cline{2-4} 
 & News & 68.2\scriptsize{$\pm$1.4} &  67.5\scriptsize{$\pm$1.6} \tabularnewline
 \cline{2-4}
 & Combined & 60.6\scriptsize{$\pm$1.4} & 60.2\scriptsize{$\pm$1.6}\tabularnewline
\hline 
\multirow{3}{*}{YouTube\scriptsize{ $\mathsf{hard}$}} & Sports & 54.4\scriptsize{$\pm$1.6} & 54.4\scriptsize{$\pm$1.6} \tabularnewline
\cline{2-4} 
& News & 69.1\scriptsize{$\pm$1.4}& 68.3\scriptsize{$\pm$1.8}\tabularnewline
\cline{2-4}
& combined & 61.5\scriptsize{$\pm$1.5} & 61.1\scriptsize{$\pm$1.8}\tabularnewline
\hline 
\end{tabular}
\end{table*}

\begin{table}
\centering
\caption{\small Category-specific experimental results on SumMe dataset. SP\_I and SP\_II stands for super-category I and II, respectively. We report 5-round results.} 
\small
\vspace{1em}
\label{tCategorySumMe}
\begin{tabular}{c|c|c|c}
Setting & Testing & \multicolumn{2}{c}{Category determined by a classifier}\tabularnewline
\hline
\multirow{2}{*}{SumMe} & SP\_I & 38.6\scriptsize{$\pm$0.5} & \multirow{2}{*}{39.2\scriptsize{$\pm$0.7}} \tabularnewline
\cline{2-3}  
 & SP\_II & 40.7\scriptsize{$\pm$0.7} &  \tabularnewline
\hline 
\multirow{2}{*}{SumMe\scriptsize{ $\mathsf{soft}$}} & SP\_I & 39.2\scriptsize{$\pm$0.6} & \multirow{2}{*}{40.2\scriptsize{$\pm$0.7}}\tabularnewline
\cline{2-3}  
 & SP\_II & 41.9\scriptsize{$\pm$0.7} & \tabularnewline
\hline 
\multirow{2}{*}{SumMe\scriptsize{ $\mathsf{hard}$}} & SP\_I & 39.8\scriptsize{$\pm$0.8} & \multirow{2}{*}{40.9\scriptsize{$\pm$0.9}}\tabularnewline
\cline{2-3} 
& SP\_II & 43.3\scriptsize{$\pm$0.9} &  \tabularnewline
\hline
\end{tabular}
\end{table}

\begin{table*}[htp]
\centering
\caption{\small Category-specific experimental results on MED summaries dataset with 10 categories. We consider two settings, $K=15$ and $K=30$, as mentioned in section~\ref{MED_data} and~\ref{MED_eval}. We report 5-round results.} 
\small
\vspace{1em}
\label{tCategoryMED}
\begin{tabular}{c|c|c}
\multicolumn{3}{c}{(a) User summaries at $K = 15$} \tabularnewline
\hline
Setting & \multicolumn{2}{|c}{Category prior not used}  \tabularnewline
\hline
MED & \multicolumn{2}{|c}{28.9\scriptsize{$\pm$0.8}} \tabularnewline
\hline
& Ground-truth category is used & Category determined by a classifier\tabularnewline
\hline 
MED\scriptsize{ $\mathsf{soft}$} & 30.7\scriptsize{$\pm$1.0} & 29.4\scriptsize{$\pm$1.2} \tabularnewline
\hline 
MED\scriptsize{ $\mathsf{hard}$} & 30.4\scriptsize{$\pm$1.0} & 28.5\scriptsize{$\pm$1.3} \tabularnewline
\hline
\multicolumn{3}{c}{} \tabularnewline
\multicolumn{3}{c}{(b) User summaries at $K = 30$} \tabularnewline
\hline
Setting & \multicolumn{2}{|c}{Category prior not used}  \tabularnewline
\hline
MED & \multicolumn{2}{|c}{47.2\scriptsize{$\pm$0.7}} \tabularnewline
\hline
& Ground-truth category is used & Category determined by a classifier\tabularnewline
\hline 
MED\scriptsize{ $\mathsf{soft}$} & 48.6\scriptsize{$\pm$0.8} & 45.9\scriptsize{$\pm$1.0} \tabularnewline
\hline 
MED\scriptsize{ $\mathsf{hard}$} & 48.1\scriptsize{$\pm$1.1} & 44.5\scriptsize{$\pm$1.1} \tabularnewline
\hline
\end{tabular}
\end{table*}

\subsection{Comparison between deep and shallow features}
\label{shallow_deep}
Here we present results on contrasting deep to shallow features to show that deep features for visual recognition do not help much over shallow features. As shown in Table~\ref{tShallowDeep}. We compare Fisher vector and color histogram with state-of-the-art CNN features by GoogLeNet \cite{szegedy2014going}.

\begin{table}
\centering
\caption{\small The comparison of shallow and deep features for summarization. We report 5-round results.} 
\small
\vspace{1em}
\label{tShallowDeep}
\begin{tabular}{c|c|c}
& {Shallow} & {Deep} \tabularnewline
\hline 
Kodak  & {\textbf{80.2}\scriptsize{$\pm$0.3}} & {79.1\scriptsize{$\pm$0.4}} \tabularnewline
\hline
{YouTube} & {\textbf{60.0}\scriptsize{$\pm$1.3}} & {59.3\scriptsize{$\pm$1.5}} \tabularnewline
 \hline 
{YouTube\scriptsize{ $\mathsf{soft}$}} & {\textbf{60.6}\scriptsize{$\pm$1.4}} & {59.6\scriptsize{$\pm$1.6}}\tabularnewline
\hline 
{YouTube\scriptsize{ $\mathsf{hard}$}} & {\textbf{61.5}\scriptsize{$\pm$1.5}} & {61.0\scriptsize{$\pm$1.6}}\tabularnewline
\hline 
\end{tabular}
\end{table}
\subsection{Comparison to seqDPP}
Our method outperforms seqDPP~\cite{gong14diverse} on Kodak and YouTube (cf. Table 2 in the main text), whose contents are more diverse and more challenging than OVP.
We suspect that since videos in OVP are edited and thus less redundant, methods with higher precision such as seqDPP might be able to perform better than methods with higher recall (such as the proposed approach).

At testing time, our model requires more computation. On YouTube, it takes 1 second on average per video, slower than 0.5 second by seqDPP. However, our model is far advantageous in training. First, it learns much fewer parameters (cf. in the main text, about 9,000 for $\valpha$ in eq.~(5) and diagonal $\mOmega$ in eq.~(4)), while seqDPP requires tuning 80,000 parameters. Our model thus learns well even on small training datasets, shown on Kodak (cf. in section~4.3 of the main text). Secondly, learning seqDPP is computationally very intensive --- according to its authors, a great deal of hyperparameter tuning and model selection was performed. Learning our model is noticeably faster.  For example, on Youtube, our model takes about 1 minute per configuration of hyperparameters, while seqDPP takes 9 minutes.

\subsection{Qualitative comparisons}

\begin{figure*}
\centering
\begin{tabular}{ll}
 & \multirow{9}{*}{\includegraphics[width=0.75\textwidth]{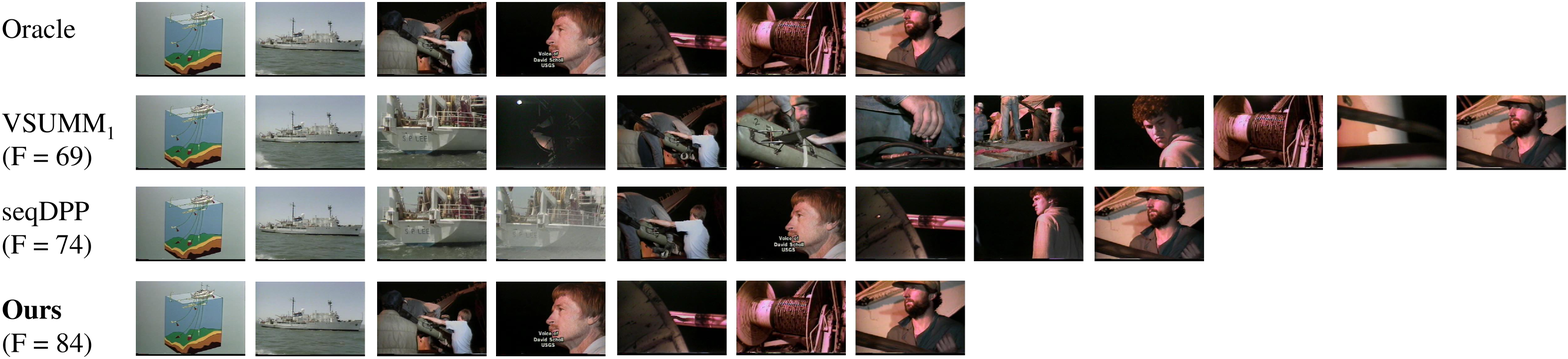}} \\
\\
\\
\small{OVP}\\
\small{(Video 60)} \\
\\
\\
\\ \hline
 & \multirow{9}{*}{\includegraphics[width=0.75\textwidth]{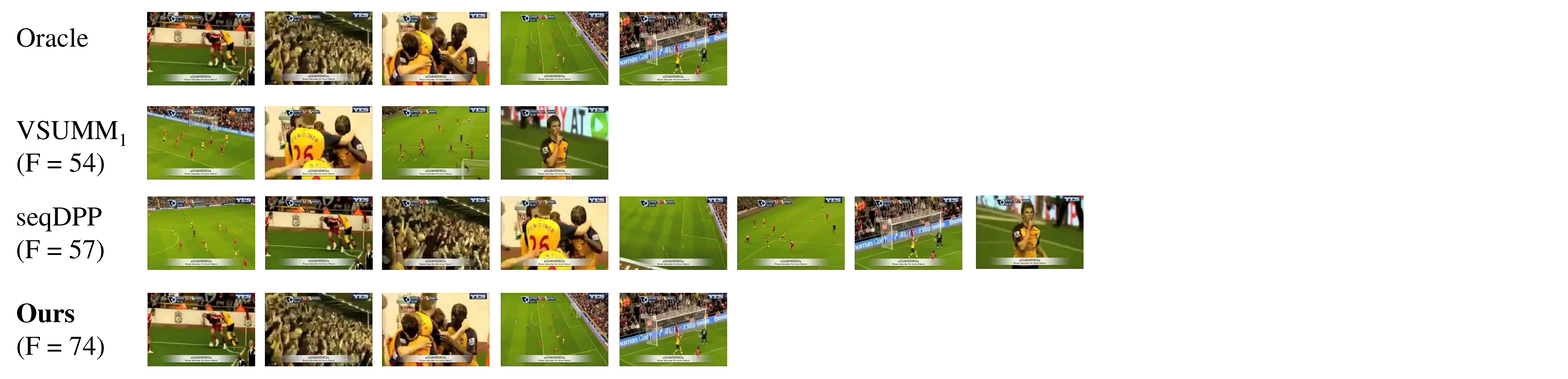}} \\
\\
\\
\small{YouTube}\\
\small{(Video 76)} \\
\\
\\
\\ \hline
& \multirow{9}{*}{\includegraphics[width=0.75\textwidth]{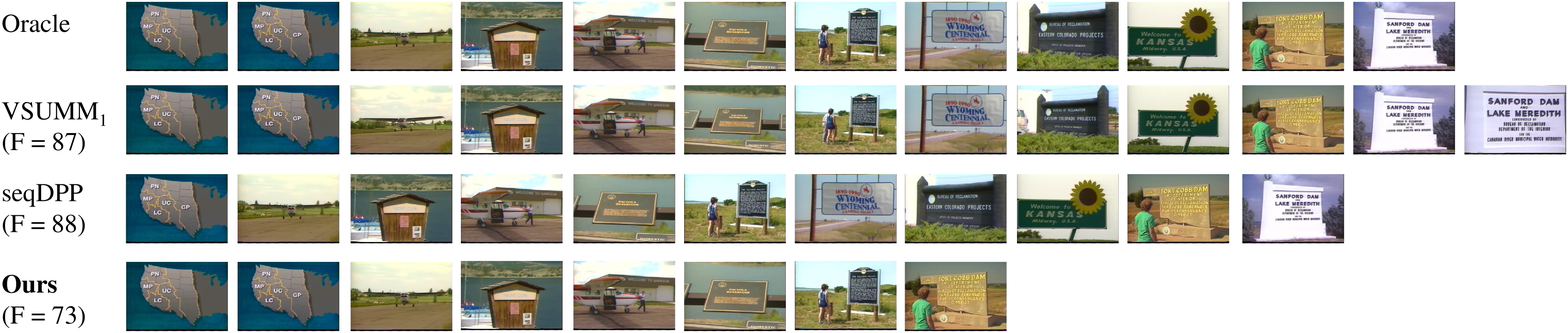}} \\
\\
\\
\small{OVP}\\
\small{(Video 25)}\\
\\
\\
\\ \hline
\end{tabular}
\vskip 1em
\caption{\small Exemplar video summaries generated by \textsc{vsumm}$_1$~\cite{de2011vsumm}, seqDPP~\cite{gong14diverse}, and ours ($\mathsf{sim3}$), along with the (merged) human-created summaries. We index videos following~\cite{de2011vsumm}. On two videos, our approach reaches the highest agreement with the oracle/human-created summaries compared to the other methods. The failure case on the bottom hints the limitation, however. See texts for details.}
\label{fQualitative}
\end{figure*}

We provide exemplar video summaries in Fig.~\ref{fQualitative} (and attached movies),   along with the human-created summaries.   Since OVP and YouTube have five human-created summaries per video, we display the merged oracle summary (see section~\ref{sDatKOY}) for brevity. 

In general, our approach provides summaries most similar to those created by humans. We attribute this to two factors. Firstly, employing supervised learning helps identify representative contents. \textsc{vsumm}$_1$~\cite{de2011vsumm}, though achieving diversity via unsupervised clustering, fails to capture such a notion of representativeness, resulting in a poor F-score and a drastically different number of summarized frames compared to the oracle/human annotations. Secondly, through non-parametrically transferring the underlying criteria of summarization, the proposed approach arrives at  better summarization kernel matrices $\mL$ than seqDPP, further eliminating several uninformative frames from the summaries (thus improving precision).  The bottom failure case, however, suggests how to improve the method further, in the direction of increasing recall rate.  We believe those missing frames (thus, shorter summaries) can be recalled if we could consider jointly the kernel computed explicitly on the test videos (such as seqDPP) and the kernel computed by our method.

\section{Detailed discussions}
\label{sdDisc}
\subsection{Computational complexity and practicality}
Section~3.5 of the main text discusses the computational cost and ways for reducing it. Specifically, the cost depends on (1) the length of the training and testing videos; (2) the number of training videos. 

For (1), we can down-sample the videos to reduce the number of frames (cf. section~3.5 of the main text and section~\ref{sDatKOY} for details), and exploit subshot-based transfer (section.~3.4 of the main text). For the latter, the size of $\mL$ (eq. (11) of the main text) depends only on the number of subshots (not frames), and the subshot-to-subshot similarity further reduces computational cost, without worsening the performance (cf.~Table 5 of the main text). Moreover, the sequential modeling trick in~\cite{gong14diverse} can also be used to reduce the cost (see section~\ref{sSeqMode}).

For (2), the proposed hard category-specific transfer (cf. section 3.4 of the main text) enables transferring from fewer training videos with improved performance.

\subsection{Applicability to long egocentric videos}
Egocentric video is challenging due to its length and diverse content (e.g., multiple events). One possible strategy is to apply existing techniques to detect and segment the videos into shorter events and then perform our approach on shorter segments, or use methods described in~\cite{lee2012discovering} to zoom into frames surrounding the occurrence of important people and objects.

\subsection{Mechanisms to check for failure}

\begin{figure}[htp]
\vskip -1em 
\centering
\includegraphics[width=0.23\textwidth]{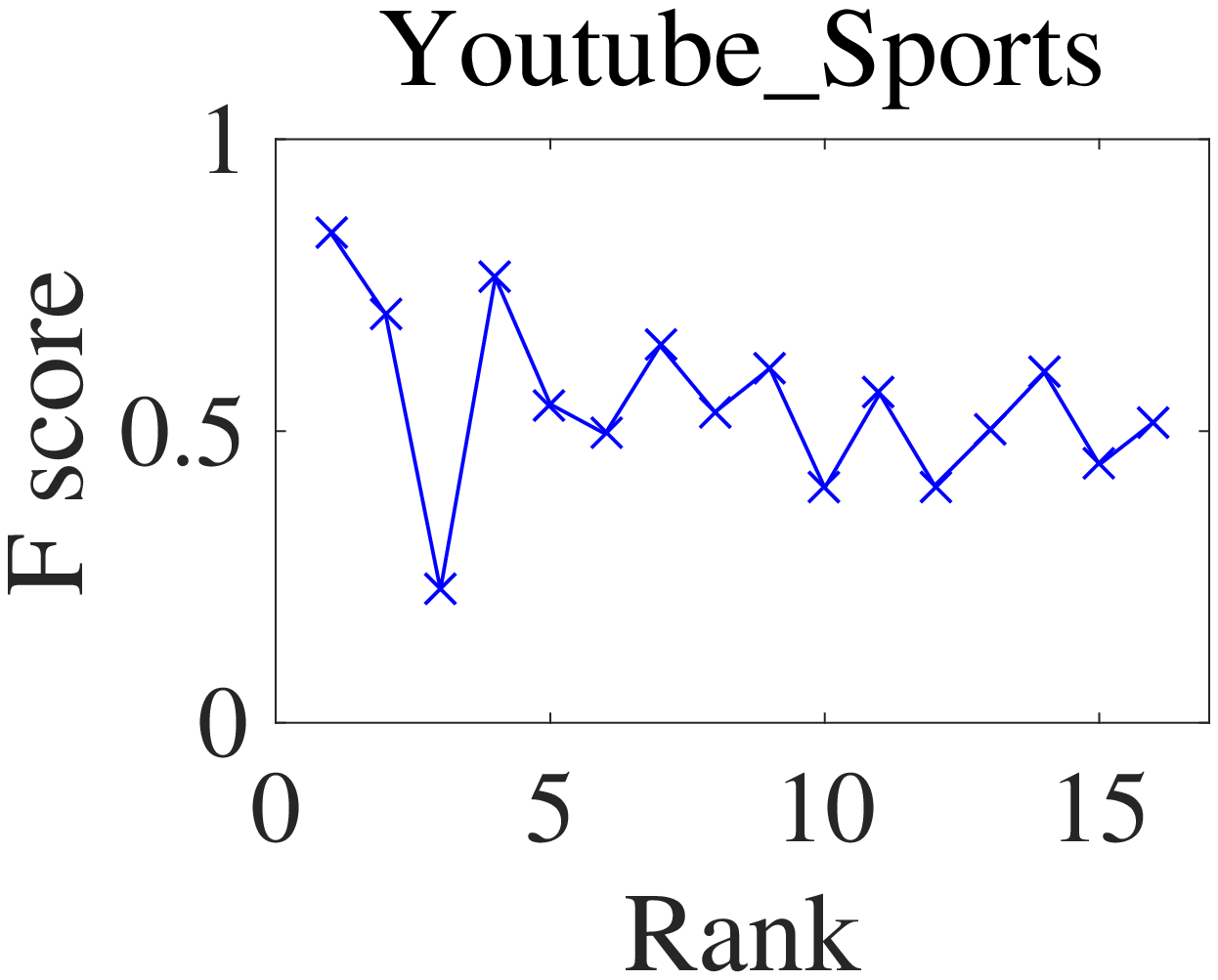}
\includegraphics[width=0.23\textwidth]{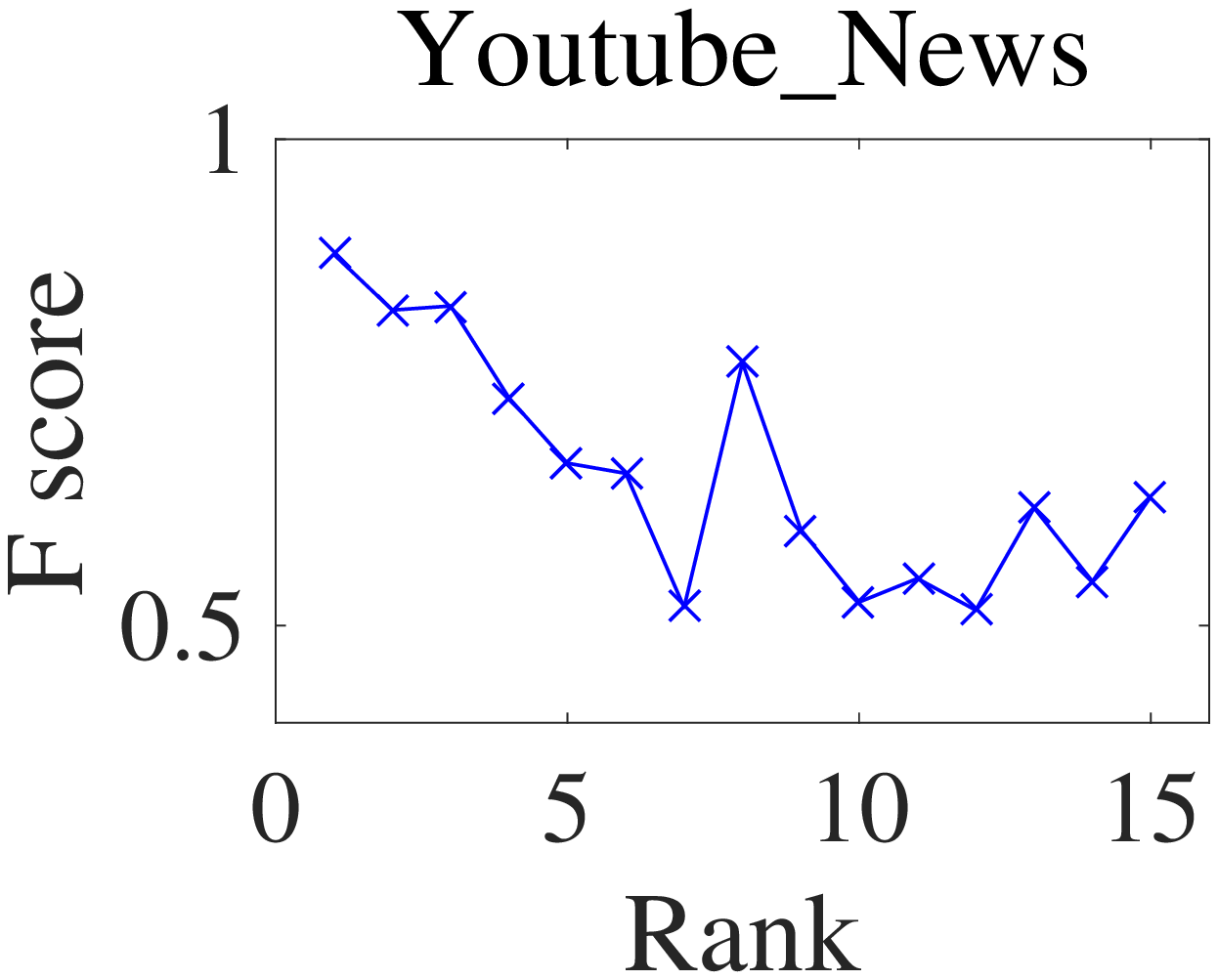}
\caption{\small Visual similarity ($x$-axis, smaller ranks imply stronger similarity) between the testing and annotated videos is positively correlated to summarization quality (F-score on $y$-axis).  Results are from summarizing YouTube via hard transfer.}
\label{fCategory_YoutubeSN}
\vskip -1em 
\end{figure}

It is interesting to investigate when our approach will fail in practice. As our approach is non-parametric by transferring the summarization structures from annotated videos, the visual similarities between the testing and annotated videos might be a useful cue.
We conduct analysis (cf. Fig.~\ref{fCategory_YoutubeSN}) and show that there is indeed positive correlation between visual similarity and summarization quality. A preliminary fail-safe mechanism thus would be thresholding on the similarity.
